\documentclass[sigconf]{acmart}
\AtBeginDocument{%
  \providecommand\BibTeX{{%
    \normalfont B\kern-0.5em{\scshape i\kern-0.25em b}\kern-0.8em\TeX}}}

\setcopyright{acmcopyright}
\copyrightyear{2018}
\acmYear{2018}
\acmDOI{10.1145/1122445.1122456}

\renewcommand\footnotetextcopyrightpermission[1]{}

\usepackage{subcaption,booktabs}

\newcommand{\qt}[1]{``#1''}

\newcommand{\modelname}{{Our model}}

\newcommand{\xhdr}[1]{\vspace{1.7mm}\noindent{{\bf #1.}}}

\graphicspath{{./fig/}}
\newcommand{\tim}[1]{\textcolor{red}{}}

\begin{document}

%%
%% The "title" command has an optional parameter,
%% allowing the author to define a "short title" to be used in page headers.
\title{Transformer-Based Behavioral Representation Learning Enables Transfer Learning for Mobile Sensing in Small Datasets}
% \title{Transformer-Based Behavioral Representation Learning Enables Transfer Learning for Small Data Mobile Sensing Time Series}
%%

%% The "author" command and its associated commands are used to define
%% the authors and their affiliations.
%% Of note is the shared affiliation of the first two authors, and the
%% "authornote" and "authornotemark" commands
%% used to denote shared contribution to the research.
\author{Mike A. Merrill}
\email{mikeam@cs.washington.edu}
\affiliation{
  \institution{University of Washington}
  \city{Seattle}
  \state{Washington}
  \country{USA}
}
\author{Tim Althoff}
\email{althoff@cs.washington.edu}
\affiliation{
  \institution{University of Washington}
  \city{Seattle}
  \state{Washington}
  \country{USA}
}

%%
%% By default, the full list of authors will be used in the page
%% headers. Often, this list is too long, and will overlap
%% other information printed in the page headers. This command allows
%% the author to define a more concise list
%% of authors' names for this purpose.
\renewcommand{\shortauthors}{Authors and Author, et al.}

%%
%% The abstract is a short summary of the work to be presented in the
%% article.
\begin{abstract}

While deep learning has revolutionized research and applications in NLP and computer vision, this has not yet been the case for behavioral modeling and behavioral health applications. This is because the domain's datasets are smaller, have heterogeneous datatypes, and  typically exhibit a large degree of missingness.
% Since these properties are uncommon in NLP and computer vision, 
Therefore, off-the-shelf deep learning models require significant, often prohibitive, adaptation. Accordingly, many research applications still rely on manually coded features with boosted tree models, sometimes with task-specific features handcrafted by experts.

Here, we address these challenges by providing a neural architecture framework for mobile sensing data that can learn generalizable feature representations from time series and demonstrates the feasibility of transfer learning on small data domains through fine tuning.
This architecture combines benefits from CNN and Transformer architectures to (1) enable better prediction performance by learning  directly from raw minute-level sensor data without the need for handcrafted features by up to 0.33  ROC AUC, and (2) use pretraining to outperform simpler neural models and boosted decision trees with data from as few a dozen participants.
\end{abstract}

%%
%% CSSXML is generated by the tool at http://dl.acm.org/ccs.cfm.
%% Please copy and paste the code instead of the example below.

%%
%% Keywords. The author(s) should pick words that accurately describe
%% the work being presented. Separate the keywords with commas.
% \keywords{some, keywords}

%% A "teaser" image appears between the author and affiliation
%% information and the body of the document, and typically spans the
%% page.
\begin{teaserfigure}
  \includegraphics[width=\textwidth]{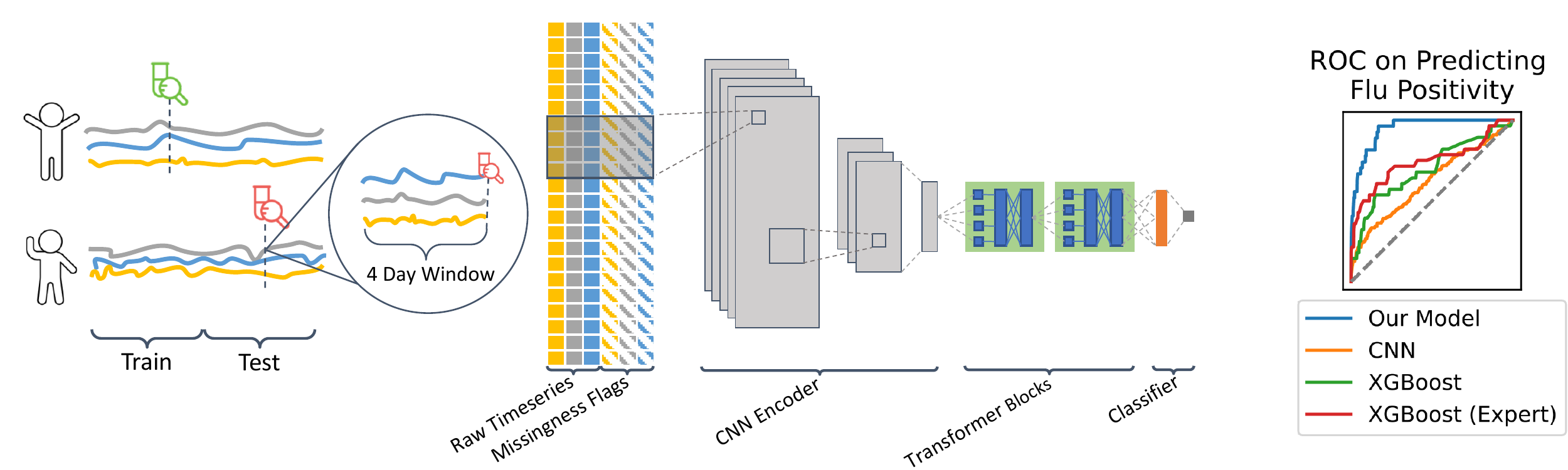}
  \caption{Our model enables transfer learning on as few as a dozen participants by applying a CNN to mine features from raw timeseries data, which are in turn passed to a transformer for classification. Our model  outperforms commonly used neural and non-neural baselines on tasks related to flu monitoring.}
  \Description{The architecture of our model, which uses a CNN on raw time series data as input to a Transformer}
  \label{fig:teaser}
\end{teaserfigure}

%%
%% This command processes the author and affiliation and title
%% information and builds the first part of the formatted document.
\maketitle
\pagestyle{plain} % removes running headers
\newcommand{\tableTaskResults}{
\begin{table}
    \begin{subtable}{\columnwidth}
    \centering
        \begin{tabular}{lllp{1.3cm}l}
            \toprule
            Task & \modelname & CNN & XGBoost (Expert) & XGBoost \\
            \hline
            Flu Positivity &  \textbf{0.94}  & 0.61 & 0.74 & 0.70 \\
            Kit Trigger & \textbf{0.69} & 0.63 & 0.68 & 0.68 \\
            Flu Symptoms & \textbf{0.68} & 0.60 & 0.62 & 0.66\\
            \hline
        \end{tabular}
        \caption{ROC AUC on each prediction task. \modelname~ outperforms all baselines, indicating that a hybrid CNN-transformer architecture is a powerful model for behavioral modeling tasks.}
        \label{tab:taskResults}
    \end{subtable}
    \begin{subtable}{\columnwidth}
        \centering
        \begin{tabular}{p{1.8cm}lp{1.4cm}l}
            \toprule
            \modelname~w/ Pretraining & \modelname & XGBoost (Expert) & XGBoost \\
            \hline
              \textbf{0.59}  & 0.51 & 0.54 & 0.54 \\
            \hline
        \end{tabular}
        \caption{ROC AUC on the Flu Symptom task after training on only twelve participants. \modelname~with pretraining outperforms baselines, indicating that pretrained transformers offer improved performance on behavioral modeling tasks with limited training data. Does this need to be a table? ROC plot won't look impressive.}
        \label{tab:pretraining}
    \end{subtable}
    \caption{\modelname's performance on behavioral modeling tasks, and its ability to be effectively finetuned in contexts with limited data show that it is a powerful tool for making inferences from time series data. Change order to match baseline description}
\end{table}
}

\newcommand{\figTaskTraining}{
\begin{figure}
    \includegraphics[width=0.95\columnwidth]{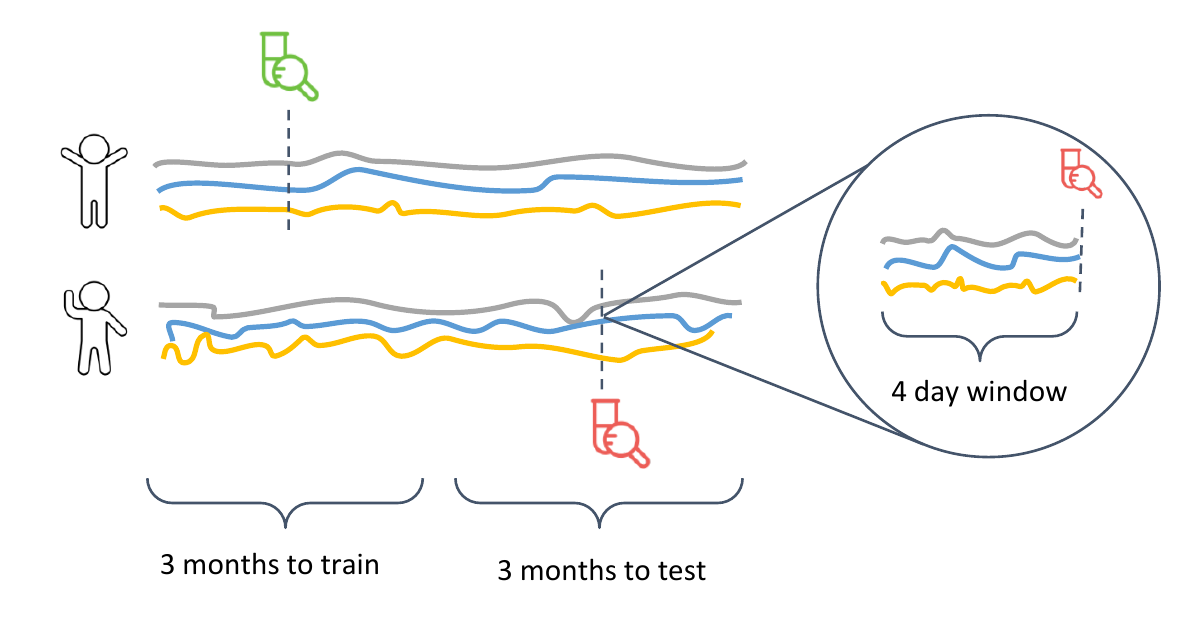}
    \caption{Model Training}
    \label{fig:task_training}
\end{figure}
}

\newcommand{\tableTaskFeatures}{
\begin{table}[h]
    \centering
    \begin{tabular}{ll}
        \textbf{Features} & \textbf{Description}\\
        \toprule
        \multicolumn{2}{l}{Standard Features}\\
        \hline
        Resting HR & Avg. heart rate (HR) while not moving\\
        Main Minutes in Bed & Longest duration of minutes in bed\\
        Sleep Efficiency & Time spent sleeping over time in bed\\
        Nap Count & Number of naps\\
        Total Asleep Minutes & Total time spent sleeping\\
        Total in Bed Minutes & Total time spent in bed\\
        Active Calories & Calories burned from exercise\\
        Calories Out & Total calories burned\\
        Base Metabolic Rate & Calories passively burned\\
        Sedentary Minutes & Time spent not moving\\
        Lightly active minutes & Time spent lightly active\\
        Fairly active minutes & Time spent lightly exercising\\
        Very active minutes & Time spent lightly exercising\\
        Missing HR & Indicator for missing HR data\\
        Missing Sleep & Indicator for missing sleep data\\
        Missing Steps & Indicator for missing steps\\
        Missing Day & Indicator for missing all data\\
        \hline
        \multicolumn{2}{l}{Expert Features}\\
        \hline
        Resting HR 95\textsuperscript{th} Pct & 95\textsuperscript{th} percentile of resting HR\\
        Resting HR 50\textsuperscript{th} Pct & 50\textsuperscript{th} percentile of resting HR\\
        Resting HR std. & Standard deviation of resting HR\\
        Awake HR 95\textsuperscript{th} Pct & 95\textsuperscript{th} percentile of HR while awake\\
        Steps Streak 95\textsuperscript{th} Pct & 95\textsuperscript{th} percentile of continuous steps\\
        Steps Streak 50\textsuperscript{th} Pct & 50\textsuperscript{th} percentile of continuous steps\\
        \bottomrule
    \end{tabular}
    \caption{Summary of manually defined features.}
    \label{tab:features}
\end{table}
}
\section{Introduction}
In recent years, deep learning has provided performance improvements across NLP and computer vision. However, most off-the-shelf methods require large datasets, which has slowed their adoption in behavioral modeling applications such as mobile sensing where researchers report difficulties collecting labeled datasets of even dozens of participants\tim{cite ORson?}. Without substantial funding for personnel and devices data collection quickly scales to an arduous process of recruiting and paying participants, debugging devices, and monitoring data quality \cite{xu2021understanding}. Therefore, researchers have historically often been limited to using small datasets, hand-crafted features, and non-neural models~\cite{xu2021understanding,laport-lopez_review_2020}.

If mobile sensing applications could achieve improved performance and generalizability with smaller datasets they could be deployed quickly in situations with limited training data. For example, in the crucial early days of a emerging disease outbreak, such as the COVID-19 pandemic, laboratory testing may not be widely available, and many positive cases may remain undetected. In this setting, a highly generalizable model could be trained on what few positive test results a researcher had, and used to identify members of a population who may be infected and should be targeted for additional testing.

In this paper we enable these conditions through transfer learning, which has recently driven advancements in NLP and  computer vision \cite{devlin2019bert,lan2019albert,mikolov2013distributed,krizhevsky2012alexnet,he2016resnet}. In each case, a researcher first obtains a model trained on a large, more readily available, task-independent dataset (such as the popular language model BERT, which was trained in part on Wikipedia) and finetunes it on a typically smaller task-specific dataset. The resulting finetuned model provides higher predictive power than either the pretrained model without finetuning or a model trained exclusively on the smaller dataset. However, existing architectures and pretraining techniques cannot trivially accept high dimensional, heterogeneous multi-modal timeseries data, and are not equipped to support missing data, both of which are rare in computer vision and NLP but endemic in behavioral modeling applications  \cite{xu2021understanding}.

Accordingly, our goal is to not only demonstrate that transformers provide significantly improved predictive power over other methods, but also how they can be coupled with pretraining to help researchers make stronger inferences with limited datasets. We first describe the HomeKit Flu monitoring study, which we use as a test bed for exploring transfer learning for behavioral data. We then show that our model provides significantly improved predictive power over other methods on three tasks related to detecting the flu. Next, we show that by pretraining our model on a large dataset and finetuning it on only one dozen participants' data we can outperform neural and non-neural baselines to deliver inferences on unlabeled participants. Finally, we discuss the implications of these findings in the context of behavioral monitoring.

\section{Related Work}
Our model builds upon prior work in neural methods and transfer learning for behavioral sensing and modeling. Our model is the first to mine raw sensor signals for generalizable feature representations to enable transfer learning in small datasets.

\subsection{Neural Models for Sensing}
Behavioral data has been modeled and mined using deep learning techniques across a variety of domains, including human activity recognition (using CNN)~\cite{yao2017deepsense},
personalized fitness recommendation (using stacked LSTM~\cite{hochreiter1997long})~\cite{ni2019modeling},
mood prediction (using RNN, GRU, or autoencoder)~\cite{suhara2017deepmood,cao2017deepmood,spathis2019sequence},
stress prediction (using LSTM and autoencoder)~\cite{li2020extraction},
and personality prediction~\cite{wu2020representation}. Two studies experimented with multi-head attention and convolution as we do here, but neither paper applies this architecture to transfer learning \cite{song2018attend,tang2021selfhar}.

\subsection{Transfer Learning and Behavioral Modeling}
Transfer learning for wearable and sensor data has been explored in human activity recognition~\cite{ma2020transfer}, stress and mood prediction~\cite{jaques2017multitask,li2020extraction}, and forecasting adverse surgical outcomes in an ICU~\cite{chen2020forecasting}. However, none of these applications focus explicitly on model performance on small datasets, as we do here. In this respect, the most relevant work is Tang et. al \cite{tang2021selfhar} which tests its methods on populations with as small as 1.5k samples. However, in this work we not only train our model on a dataset with less than half as many samples, but also show that its predictions can be generalized to a population up to five hundred times as large (Section \ref{subsec:results_transfer_learning}). 
\section{Methods}
We first describe a dataset of FitBit recordings and flu test results, which we use as a test bed for transformer-based behavioral representation learning. Then, we detail our model which is composed of a CNN Encoder for capturing hierarchical and temporal features and transformer for learning relationships between these features.
\subsection{Dataset}
\tableTaskFeatures

\xhdr{Data and Hand-Crafted Features} Our dataset consists of 118k user-days of FitBit data collected from 983 participants in the Homekit Flu Monitoring Study over the course of six months. Each minute the devices recorded the participant's total steps, average heart rate, and a binary flag indicating if the participant was sleeping. Participants also completed daily surveys which asked if they were experiencing flu symptoms, including coughing, chills, fever, and fatigue. When a participant indicated that they were experiencing a cough and one other symptom, they were asked to self-administer a saliva swab test kit, which was then mailed to a lab for further analysis. 

\xhdr{Missing Data}
We note that our dataset demonstrates modest missingness, with the median participant supplying data on 114 of 120 possible days. Researchers frequently report missingness as an obstacle to adopting deep learning techniques, and so we model missingness by replacing missing values with zeros and including a binary flag at each timestep for each sensor stream to indicate if the value is missing in that timestep.

\subsection{Model}
\modelname~is composed of a convolutional encoder, a stack of transformer blocks, and a final densely connected linear layer that is used for classification. Intuitively, the convolutional encoder learns a compressed, hierarchical feature representation of the raw timeseries data, while the transformer learns relationships between these features. An overview of our architecture is available in Figure \ref{fig:teaser}. 

\xhdr{Convolutional Encoder} The convolutional encoder learns a temporal, hierarchical feature representation of the raw sensor data. We experimented with several architectures, and found that three layers with kernel sizes of 5,3,1, stride sizes of 2,2,2, and output channels of 256,128,64 worked best, although in practice the model does not appear to be particularly sensitive to this module's hyperparameters. Note that unlike SAnD (which applies convolution across the features at each timestep), we treat each sensor stream as an input channel, and compute multiple convolutions between timesteps \cite{song2018attend}. As a result, the input to our transformer blocks is significantly compressed. We experimented with SAnD's convolutional layer, but failed to achieve above random performance on any task. We suspect that this is because our model's inputs are longer (four days of data as opposed to one) and \qt{thinner} (six features as opposed to 76), meaning that compression along the feature axis alone does little to compress the overall dimensionality.

\xhdr{Transformer Blocks} Our model uses a stack of two transformer layers, each composed of four attention heads and a feed forward layer. We take the output of the final layer to be the learned representation of the input time series. 

\xhdr{Training} We train our model with the Adam optimizer \cite{kingma2017adam} and Focal loss \cite{lin_focal_nodate}, which has been previously shown to be highly effective in cases such as ours that exhibit extreme class imbalance (roughly 500:1 in the case of the Kit Trigger task). We selected data from a subset of 10\% of users in our test set, and used this data as an evaluation set for hyperparameter tuning. 

\tableTaskResults
\section{Results}
To evaluate our model's performance we simulate a realistic setting where a hypothetical researcher deploys a model's predictions on a population after an initial training period (Figure \ref{fig:teaser}). We train on the first three months of data, test on the following three months, and make a prediction on each task for every participant on every day of the study. Additionally, our model only uses data from the four days prior to the day on which we make a prediction, so that no information from the future is used to make a prediction about the past. We also include no explicit information about a users identity (e.g. participant id or demographics) to encourage the model to learn generalizable motifs about activity data rather than facets of individual users' behavior.   
This evaluation setting follows best-practice recommendations and avoids falsely overstating the level of performance~\cite{nestor2021dear}.

\subsection{Single Domain Prediction Tasks}
\label{subsec:pred_tasks}
We evaluate our model on three behavioral modeling tasks: 

\begin{itemize}
    \item \textbf{Flu Positivity:} Will the participant produce a swab that tests positive for the flu today?
    \item \textbf{Kit Trigger:} Will the participant trigger their home test kit today (i.e., report a cough and at least one other symptom)?
    \item \textbf{Flu Symptoms:} Will the participant report any flu symptoms today?
\end{itemize}

In each case, we compare our model to the following baselines:
\begin{itemize}
    \item \textbf{CNN:} How important are the transformer layers to our pretrained model's performance? To answer this question, we removed the transformer blocks from our model and passed the CNN's final output directly to a linear layer. 
    \item \textbf{XGBoost:} How well does our model perform relative to a non-neural baseline? Boosted decision trees are frequently used in many sensing studies because they are supported by common, easy to use libraries and often achieve strong performance out-of-the-box \cite{xu2021understanding}. Since boosted trees expectedly do not scale well to the thousands of observations in our raw time series data, we compute a set of commonly used features for each day in the window, and then concatenate these features for a final input. A list of all features is available in Table \ref{tab:features}. 
    \item \textbf{XGBoost - Expert:} What if our non-neural model had access to features that were designed by experts? This model is similar to the previous baseline, but we add six features from prior work which are shown to be relevant to respiratory viral infections \cite{radin_assessment_2021}.
\end{itemize}

We do not include a \qt{transformer only} baseline (i.e. our model but without the CNN encoder) because multi-head attention scales quadratically in memory with the length of the input, making it computationally infeasible to perform such an experiment on a multi-day timeseries window (i.e. minute level data on a four day window produces a 5,760 dimensional vector, which exceeds common context sizes in transformer models for NLP) with commodity GPUs \cite{beltagy_longformer_2020}.

\xhdr{Results} Our model outperforms all baselines on all three tasks (Table \ref{tab:taskResults}). Specifically, we perform 13\% to 53\% better than a CNN alone, and 1.4\% to 27\% better than XGBoost models. Interestingly, our model's marginal performance gain on the Flu Positivity task is much higher than on other tasks. One might expect that predicting flu positivity is inherently more difficult than predicting symptoms alone (as in the other two tasks) since the former requires the model to separate participants who are sick with some other respiratory viral infection from flu positive participants. However, we theorize that participants who are flu positive are more likely to display the kind of behavioral changes (e.g. staying home from work, sleeping in later) that our model is designed to capture.

\subsection{Transfer Learning with Small Datasets}
\label{subsec:results_transfer_learning}

What if a researcher was interested in making inferences about their whole population, but only had access to labels from a small subset of the population (as might be the case in an emerging disease scenario)? Here we showcase our model's  ability to learn from small datasets with pretraining and finetuning. 

\xhdr{Pretraining} The HomeKit Flu monitoring study included a daily questionnaire which asked participants to indicate if they were experiencing fatigue. Much the same way as Section \ref{subsec:pred_tasks}, we take this response as a label and associate it with a four day window of data preceding the response. We then pretrain our model on the first three months of data from the study. This process simulates a researcher using all available data up to the point where they begin testing for a new condition, like a novel disease.

\xhdr{Finetuning}
Next, we randomly select twelve participants and then use their data (constituting 600 user-days) from the second three month period to finetune the model on the \qt{Flu Symptoms} task (Section \ref{subsec:pred_tasks}). This small dataset size mirrors many projects in ubiquitous computing, mobile sensing, and clinical studies, which use on the order of a dozen participants for their inferences \cite{low_digital_2021,wang2016crosscheck,tseng_digital_2021}.  Finally, we used this finetuned model to make predictions about the remaining participants in the second three months of data. This step simulates a researcher using limited information from a small subset of their population to make inferences about the remaining participants.

\xhdr{Results}
With training data from just twelve users, our pretrained model outperforms all baselines on the Flu Symptoms task (Table \ref{tab:pretraining}). Notably, our model with pretraining  outperforms XGBoost on this task (0.59 ROC AUC v.s. 0.54 ROC AUC) while the same model \emph{without} pretraining barely performs above random chance (0.51 ROC AUC). This result shows that our model could be deployed in scenarios with limited information from which to draw inferences. 
\section{Discussion}
In this paper we presented the first transformer-based architecture for behavioral representation learning that can learn directly from raw sensor data and, when coupled with pretraining-based transfer learning, can be effectively trained on as few as a dozen users' data. We believe that a model like ours could feasibly transform behavioral modeling by providing a common generalizable set of learned feature representations for small data applications across domains like mobile sensing, ubiquitous computing, and machine learning for health. Such a tool could democratize behavioral data science by offering significantly improved predictive performance to researchers who lack the resources or opportunity to scale data collection or model training beyond dozens of participants and extend recent performance advances in NLP and computer vision to behavioral modeling.

\tim{if you have time, clean up the references as we usually do, but don't sweat it}

\bibliographystyle{ACM-Reference-Format} 
% \balance
\bibliography{references}
\end{document}